\documentclass[journal]{IEEEtran} 
\IEEEoverridecommandlockouts
\usepackage{cite}
\usepackage{svg}
\usepackage{amsmath,amssymb,amsfonts}
\usepackage{algorithmic}
\usepackage{graphicx}
\usepackage{textcomp}
\usepackage{xcolor}
\usepackage{multirow}
\usepackage{subcaption}
\usepackage{comment}
\def\BibTeX{{\rm B\kern-.05em{\sc i\kern-.025em b}\kern-.08em
    T\kern-.1667em\lower.7ex\hbox{E}\kern-.125emX}}
\begin{document}

\title{TSKAN: Interpretable Machine Learning for QoE modeling over Time Series Data}
\vspace{-2em}
\author{
\IEEEauthorblockN{Kamal Singh$^{1}$, Priyanka Rawat$^{2}$, Sami Marouani$^{1}$, Baptiste Jeudy$^{1}$}\\
\IEEEauthorblockA{
\small
\textit{$^{1}$Université Jean Monnet Saint-Étienne, CNRS, Inst. d’Optique Graduate School, Lab. Hubert Curien, F-42023 Saint-Étienne, France} \\
\textit{$^{2}$Université Jean Monnet Saint-Étienne, Telecom Saint-Etienne, F-42023 Saint-Étienne, France}\\
Email: \{firstname.surname\}@univ-st-etienne.fr}
\vspace{-2em}}

\maketitle
\thispagestyle{empty}  
\begin{abstract}
Quality of Experience (QoE) modeling is crucial for optimizing video streaming services to capture the complex relationships between different features and user experience. We propose a novel approach to QoE modeling in video 
streaming applications using interpretable Machine Learning (ML) techniques over raw time series data.
Unlike traditional black-box approaches, our method combines Kolmogorov-Arnold Networks (KANs) as an 
interpretable readout on top of compact frequency-domain features, allowing us to capture 
temporal information while retaining a transparent and explainable 
model. We evaluate our method on popular datasets and demonstrate its enhanced accuracy in QoE prediction, while offering transparency and interpretability.
  
\end{abstract}

\begin{IEEEkeywords}
    QoE, DASH, Adaptive HTTP streaming, Interpretable machine learning, KAN, Time-series
\end{IEEEkeywords}

\section{Introduction}
Quality of Experience (QoE) is a crucial aspect in today's digital landscape, as it directly affects how users perceive and interact with applications and services. Defined as ‘the overall acceptability of an application or service, as perceived subjectively by the end user’ \cite{itu-t-sg12}, QoE is a complex and multifaceted concept that cannot be solely defined by traditional Quality of Service (QoS) metrics such as bandwidth, delay, or jitter.

In the context of video streaming, for instance, perceptual quality is a critical component of QoE. While objective assessment tools can provide insights into technical aspects of video quality, they often fail to accurately capture human-perceived video quality. Subjective 
quality assessment methods, on the other hand, are time-consuming and expensive.

This paper addresses the problem of QoE modeling and prediction in video streaming applications, which is crucial for their design and optimization. While numerous QoE models have been proposed, many rely on expert knowledge. In contrast, we take a different approach by focusing on Machine Learning (ML)-based models, which offer greater potential for scalability and automation.

However, even ML-based models can fall short when relying solely on black-box deep learning approaches. These models may provide impressive accuracy but struggle to explain why certain predictions are made, making it challenging to identify the underlying factors that influence 
QoE. To bridge this gap, we propose the use of interpretable Machine Learning (ML) for QoE estimation in video streaming applications.
This enables us to not only predict QoE with high accuracy but also provide insights into the key factors that impact user experience, ultimately driving 
more informed decision-making in the development and optimization of video streaming services. 

Modeling QoE from streaming logs is inherently a multivariate time-series problem with long-range and cross-scale dependencies, and a nonlinear, order-dependent mapping from events (e.g., bitrate shifts, rebuffering) to user experience. Although black-box encoders such as CNNs, LSTMs, and Transformers can capture these effects, their internal workings remain opaque, making the models harder to interpret and trust. To couple accuracy with transparency, we use Kolmogorov–Arnold Networks (KANs)~\cite{liu2024kan} to place interpretability at the core: their spline-based univariate components make marginal effects and interactions visible and amenable to simplification. However, KANs are not immediately suited to time series; treating every timestep independently discards temporal structure and scales parameters poorly. We resolve this by combining temporal dynamics via compact frequency-domain features and using KANs as an interpretable readout on top, preserving temporal information while retaining a transparent model.

The contributions of this paper are summarized as follows:
\begin{itemize}
    \item Interpretable machine learning approach for Time-Series: We propose the combination of Kolmogorov–Arnold Networks (KANs) as an interpretable readout on top of compact  frequency-domain features, allowing for the preservation of temporal information while being interpretable.
    \item Quality of Experience (QoE) modeling and prediction: We use above interpretable machine learning approach for QoE modeling over raw time series data, providing a transparent and explainable model that can help streaming services optimize their QoE for users.

\end{itemize}

This paper is organised as follows. Section~\ref{sec:related} discusses background and related works. Section~\ref{sec:tskan} presents the interpretable machine learning architecture, which is used in section~\ref{sec:qoe} for QoE modelling. Section~\ref{sec:results} compares different approaches and finally Section~\ref{sec:conclusion} concludes the paper.

\section{Related works}
\label{sec:related}

\subsection{Interpretable Machine Learning}
 Recently, there has been a growing recognition that black-box models can 
be opaque and difficult to understand, leading to a lack of trust in their outputs. Interpretability~\cite{molnar2020interpretable, du2019techniques} has become an increasingly important aspect of machine learning, particularly in domains like networking where model decisions have significant consequences for users~\cite{guo2020explainable}. 
To address this issue, researchers have developed various techniques to make machine learning models more interpretable. Some common approaches include model-agnostic explanations. These methods, such as LIME~\cite{ribeiro2016should}, SHAP~\cite{lundberg2017unified}, etc., provide feature importance scores or other forms of explanations that are independent of the specific model being used. 

For time series data, deep learning with new architectures and techniques has introduced models such as Transformers~\cite{vaswani2017attention, zerveas2021transformer}, LSTMs~\cite{hochreiter1997long}, Fully Convolutional Networks (FCNs)~\cite{wang2017time}, ResNet~\cite{wang2017time},  InceptionTime~\cite{ismail2020inceptiontime}, etc. However, interpreting models for time series data adds yet another layer of complexity as now temporal patterns and dynamics need to be explained as well.  Some recent works like AIChronoLens~\cite{fiandrino2024aichronolens} address this problem by using SHAP for time-series.

Compared to the above post-hoc approaches, another category is to use inherently interpretable AI frameworks such as decision trees, linear regression, etc. Our proposed approach falls into this second category.  We advance the state of the art by proposing \emph{inherently interpretable} model that is specifically designed for time series data, while showing improved performance.

\subsection{QoE estimation}
Different works in QoE modeling for Dynamic Adaptive HTTP Streaming (DASH) have focused on developing models that can accurately predict the impact of various parameters on user experience \cite{singh2012quality}. An overview of various QoE models, influence factors, and subjective test methods is provided in~\cite{barman2019qoe}.
Several works have focused on feature-based prediction of streaming video. For example, Video ATLAS \cite{bampis2018feature} combines QoE-related features, including 
distortions, stalling, and memory features to make predictions. The approach in \cite{bampis2018recurrent} addresses time-varying QoE prediction. Additionally, the work in 
\cite{bampis2018towards} improves perceptually-optimized adaptive video streaming by designing a QoE database that incorporates perceptual video quality principles into 
different stages of a streaming system. Past works also include standards such as ITU-T P.1203.1 (P.NATS)~\footnote{https://www.itu.int/rec/T-REC-P.1203-201710-I/en}. 

The work in  \cite{tran2020overall} used Long Short-Term Memory (LSTM) network and linear regression module for overall quality prediction. Furthermore, \cite{barman2019no} presents two no-reference machine learning-based quality estimation models for gaming video streaming using features such as bitrate, resolution, and temporal information. However, these models are limited to compression and scaling artifacts, 
and their performance on videos encoded with other codecs needs to be explored. 

Some works have focused on incorporating prior knowledge of the human visual system and human-annotated data. For example, 
\cite{duanmu2019knowledge} presents an integrated model that includes these factors. However, this model requires further improvement to account for features like motion 
strength, re-buffering, and quality adaptation.

Other researchers have proposed cumulative quality models for DASH, such as the one introduced in \cite{tran2021cumulative}. Additionally, the concept of edge-C3 has gained 
attention for improving network resource utilization and QoE for end users. In this context, \cite{jedari2020video} discusses video edge-C3 challenges and solutions in 
next-generation wireless and mobile networks.

Most existing works on QoE modeling require expert-crafted features, which can be a limiting factor for non-experts or when dealing with complex systems. Additionally, some machine learning-based QoE modeling are black box approaches. 
As compared to above approaches, we propose an interpretable machine learning approach that operates directly on 
raw time series data, eliminating the need for expert features. We compare our approach to 
previous work on explainable QoE modeling such as \cite{wehner2023explainable}, and show improved performance.

\section{Kolmogorov-Arnold Networks for Timeseries}
\label{sec:tskan}

\begin{figure}[h!]
    \centering
    \includesvg[width=0.6\linewidth]{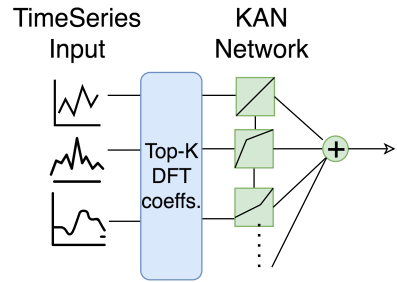}
    \caption{TSKAN architecture}
    \label{fig:tskan}
\end{figure}

We propose Kolmogorov-Arnold Networks for Timeseries (TSKAN), an interpretable machine learning method for time series regression. TSKAN leverages frequency-domain features to provide a compact representation of temporal dynamics.  TSKAN combines discrete Fourier transform (DFT) with Kolmogorov-Arnold Networks (KANs)~\cite{liu2024kan}, as shown in Figure~\ref{fig:tskan}. DFT captures periodic temporal patterns that may not be apparent in the raw time domain, while KAN learns non-linear functions of different variables and frequency components while being interpretable.

KAN operationalizes the Kolmogorov-Arnold representation by learning univariate components as smooth splines with 
regularization. This allows for interpretable component functions, which can be visualized and 
regularized for smoothness or sparsity.

From the Kolmogorov–Arnold representation theorem, any continuous map \(f:\mathbb{R}^n\!\to\!\mathbb{R}\) admits a decomposition
\begin{equation}
f(u_1,\ldots,u_n)=\sum_{q=1}^{2n+1}\Phi_q\!\Bigl(\sum_{p=1}^n\psi_{q,p}(u_p)\Bigr),
\label{eq:ka}
\end{equation}
where \(\psi_{q,p}\) and \(\Phi_q\) are univariate functions. KANs operationalize \eqref{eq:ka} by \emph{learning} these univariate components as smooth splines with regularization, thus replacing opaque weight matrices of a classic neural network with a bank of one-dimensional, transparent activations \cite{liu2024kan}. For input \(\mathbf{u}\in\mathbb{R}^{m_0}\) and \(m_1\) output channels, a KAN layer computes
\begin{equation}
o_q=\sum_{p=1}^{m_0}\psi_{q,p}(u_p),\qquad h_q=\Phi_q(o_q),\quad q=1,\dots,m_1,
\label{eq:kan-layer}
\end{equation}
where the model parameters are themselves interpretable functions (\(\psi_{q,p}\), \(\Phi_q\)) rather than fixed nonlinearities.

Existing research shows that KANs attain competitive predictive performance while providing transparent component functions that can be visualized, regularized for smoothness/sparsity, and pruned for compactness \cite{liu2024kan,somvanshi2024survey,xu2024kolmogorov}. A recent survey consolidates practical design choices of splines (knot placement, basis size, curvature control) and documents applications across tabular regression and scientific surrogate modeling, highlighting the role of KANs as interpretable universal approximators \cite{somvanshi2024survey}. 

For QoE modeling, these properties are particularly beneficial. First, the learned univariate functions make domain-relevant nonlinearities explicit: convex penalties for rebuffering, and asymmetries between up- and down-switches are directly visible in the shapes of \(\psi_{q,p}\) and \(\Phi_q\). 
Second, when driven by compact frequency-domain summaries of a video session, the resulting KAN models remain small and data-efficient, which is advantageous under sparse, noisy subjective labels and across datasets. These considerations motivate our use of a KAN-based readout for interpretable QoE estimation.
\subsection{Multivariate Time Series Representation}
We consider a multivariate time series dataset
\[
X \in \mathbb{R}^{N \times V \times T},
\]
where $N$ is the number of samples, $V$ is the number of variables (channels), and $T$ is the number of time steps per sample.  
Each sample is denoted as
\[
x^{(n)} = \big( x^{(n)}_{v}(t) \big)_{v=1,\dots,V; \, t=0,\dots,T-1}.
\]

\subsection{Frequency-Domain Feature Extraction}
For each variable $v$, we compute the discrete Fourier transform (DFT) over the temporal dimension:
\[
X_v(f) = \sum_{t=0}^{T-1} x_v(t) \, e^{-j 2\pi ft / T}, \quad f=0,\dots,T-1.
\]

From $X_v(f)$, we extract:
\begin{itemize}
    \item \textbf{DC component:} $X_v(0)$, representing the average level,
    \item \textbf{Magnitude:} 
    \[
    M_v(f) = |X_v(f)|,
    \]
    \item \textbf{Phase:} 
    \[
    \phi_v(f) = \arg(X_v(f)).
    \]
\end{itemize}

In first iteration, we retain the first $F$ frequencies ($f=0,\dots,F$) for each variable, resulting in a frequency-domain feature vector:
\[
f^{(n)} = \big[ M_v(f), \phi_v(f) \;|\; v=1,\dots,V;\; f \in \mathcal{F} \big],
\]
where $\mathcal{F}$ is the set of retained frequencies (next we will retain only top-$k$).

\subsection{Top-K frequency domain features fed into KAN}

In next iteration, we extract the most relevant frequency components.
The selection method is detailed as follows. First, the TSKAN model is trained using the $F$ frequency components, including their phases, fed to a single layer of Kolmogorov–Arnold Networks (KAN). Then, we sort them by their contribution to the output by computing the input importance scores $\alpha_j$ for each $j = 1, \ldots, F$. Finally, only the top-$k$ frequency components are retained.
Thus, $k$ and $F$ are treated as hyperparameters that can be optimized as per the use case. By selecting the top-$k$ features with highest importance scores, we reduce the number of inputs to the KAN and achieve better interpretability (as too many features reduce interpretability).

\section{QoE modeling using TSKAN}
\label{sec:qoe}

We employ TSKAN for QoE modeling over time series data, focusing on Dynamic Adaptive HTTP Streaming (DASH), a widely adopted technology in web-based applications. 
DASH segments media files into small chunks with durations of a few seconds, enabling adaptive streaming by allowing each chunk to be decoded independently. This independence facilitates seamless switching between quality levels when network conditions change.

While DASH does not suffer from video data loss due to its reliable transport layer, it may still experience playout interruptions, which can be annoying for users and must be considered in QoE estimation. Additionally, when adaptive bit-rate video streaming is used, the video quality can vary due to compression, making quantization an essential factor. To adapt to bandwidth fluctuations, the client switches to different quality. For example, the client might request low-bitrate video chunks with increased compression, which introduces distortion and impacts QoE.
For QoE modeling, we discuss the raw features that 
significantly influence user satisfaction as follows:

{\bf Stalling (seconds):} By stalling or playout interruptions, we refer to pauses in the playout without any skip of video data. Video can be paused in case when some immediate chunks are not yet downloaded. 

{\bf Bitrate (kbps):} Video bitrate impacts QoE. Reducing the bitrate typically results in inferior perceptual quality, making it essential to balance bitrate and quality for optimal viewing experiences. It is tied to QP as discussed below.

{\bf Chunk Size (seconds):} Surprisingly, chunk duration was found to impact QoE. It is determined by scene cuts. Thus, using timeseries data can capture content complexity to some extent.

{\bf Quantization Parameter (QP):} QP is another important factor affecting QoE. It determines the compression loss. In different codecs such as H.264, the value of QP can vary over time and across frames or macro blocks to achieve the target video bitrate for encoded chunks.

{\bf Frame Rate (fps) and Resolution (pixels):} In general, a higher frame rate (images per second) or image resolution are associated with better QoE. This is because a higher frame rate and resolution allow for smoother motion and more detailed visuals, enhancing the overall viewing experience.
However, they also increase bandwidth requirements, 
highlighting the need to balance frame rate and resolution with bitrate and other factors to optimize QoE.

We use Mean Opinion Score ratings (MOS) as a measure of QoE. We train TSKAN to predict MOS using time series of the above features as input. Subsequently, when making predictions, we feed the input timeseries data into TSKAN, which estimates MOS while being interpretable.

\begin{figure*}
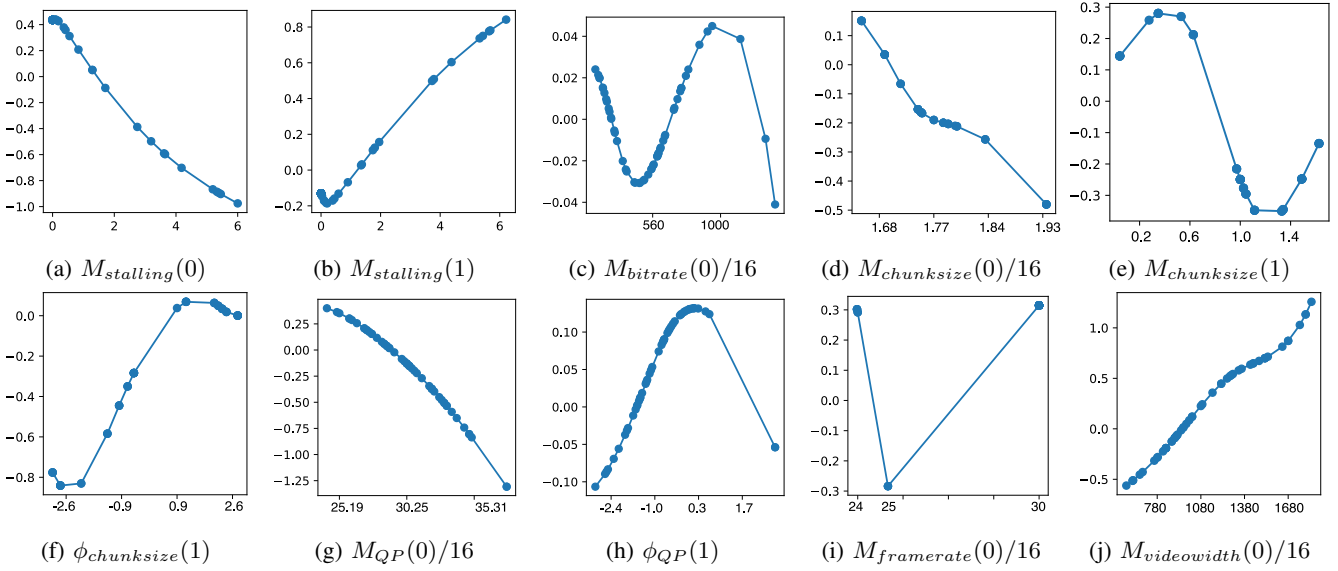

    \centering
    \begin{subfigure}{0.19\textwidth}
        \includesvg[width=\linewidth]{figures/fun0}
        \caption{$M_{stalling}(0)$}
    \end{subfigure}
    \begin{subfigure}{0.19\textwidth}
        \includesvg[width=\linewidth]{figures/fun1}
        \caption{$M_{stalling}(1)$}
    \end{subfigure}
    \begin{subfigure}{0.19\textwidth}
        \includesvg[width=\linewidth]{figures/fun2}
        \caption{$M_{bitrate}(0)/16$}
    \end{subfigure}
    \begin{subfigure}{0.19\textwidth}
        \includesvg[width=\linewidth]{figures/fun3}
        \caption{$M_{chunksize}(0)/16$}
    \end{subfigure}
    \begin{subfigure}{0.19\textwidth}
        \includesvg[width=\linewidth]{figures/fun4}
        \caption{$M_{chunksize}(1)$}
    \end{subfigure}

    \begin{subfigure}{0.19\textwidth}
        \includesvg[width=\linewidth]{figures/fun5}
        \caption{$\phi_{chunksize}(1)$}
    \end{subfigure}
    \begin{subfigure}{0.19\textwidth}
        \includesvg[width=\linewidth]{figures/fun6}
        \caption{$M_{QP}(0)/16$}
    \end{subfigure}
    \begin{subfigure}{0.19\textwidth}
        \includesvg[width=\linewidth]{figures/fun7}
        \caption{$\phi_{QP}(1)$}
    \end{subfigure}
    \begin{subfigure}{0.19\textwidth}
        \includesvg[width=\linewidth]{figures/fun8}
        \caption{$M_{framerate}(0)/16$}
    \end{subfigure}
    \begin{subfigure}{0.19\textwidth}
        \includesvg[width=\linewidth]{figures/fun9}
        \caption{$M_{videowidth}(0)/16$}
    \end{subfigure}
    \caption{
    We show top 10 activation functions  of frequency domain features after training on LIVE-netflix-II. For a feature $v$, $M_v(0)$ is DC component which is a sum of 16 values. Thus, sometimes we show the divided features by 16 to project them around comprehensible values like framerate of 24 fps instead of the feature value of 24x16. The unit is same as corresponding feature described in Section~\ref{sec:qoe}. $M_v(1)$ is magnitude of frequency $k=1$, with units as described before, $\phi_{v}(1)$ is the phase of frequency $k=1$ and it varies from $-\pi$ to $\pi$, with units in radians. }
    \label{fig:allfun}
\end{figure*}

\begin{figure}
    \centering
    \includesvg[width=1\linewidth]{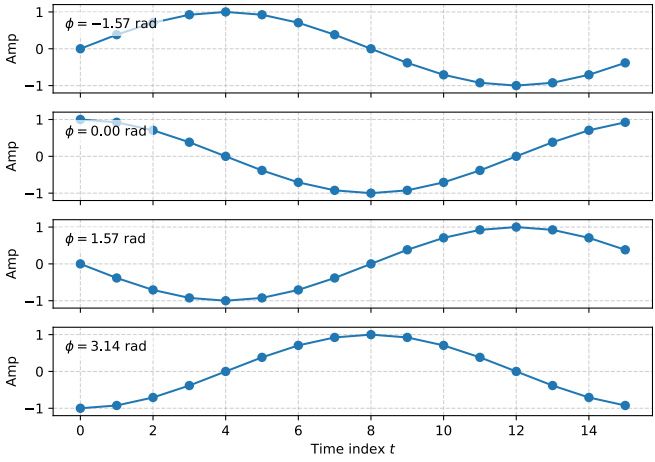}
    \caption{Illustration of cosine wave shapes for f = 1 with different phase values.}
    \label{fig:phase}
\end{figure}

\begin{figure}
    \centering
    \includesvg[width=1\linewidth]{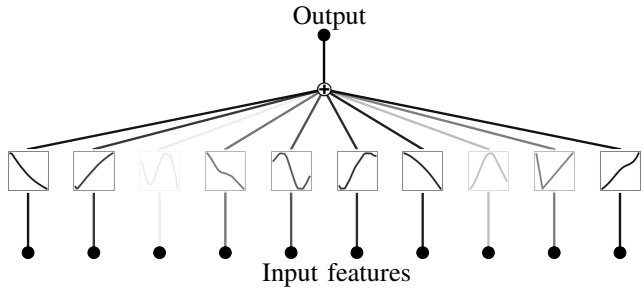}
    \caption{Output of 1-layer KAN is the sum of trained activations of selected frequency-domain features. Importance is shown using intensity.}
    \label{fig:kan}
\end{figure}

\section{Performance Evaluation}
\label{sec:results}

Experiments were conducted on a machine equipped with a Xeon CPU, 32 GB RAM, and an NVIDIA A2000 GPU (12 GB), using the \texttt{PyTorch} and \texttt{scikit-learn} libraries.
We evaluated our approach using 2 datasets: Waterloo III dataset~\cite{duanmu2018quality} and LIVE-Netflix-II dataset \cite{bampis2018towards}. Videos are divided into chunks, where each chunk corresponds to one time step. In above datasets, we have a sequence length of 5 or 16 time steps, respectively.
Note that the LIVE-Netflix-II dataset has variable length data. To ensure a consistent number of chunks, we removed samples with more than 16 chunks, as there were only 29 such examples. The quality scores were obtained by first performing z-scoring on the per-subject ratings and then averaging across subjects. The resulting z-scored MOS (Mean Opinion Score) values range from $-2.5$ to $2.5$. The data were split into 70\% training, 15\% validation, and 15\% test sets. All features were scaled using the \texttt{RobustScaler} from \texttt{scikit-learn}. 
We compare TSKAN with black-box deep learning models and other explainable QoE modeling work \cite{wehner2023explainable} as well as some inherently explainable models, as described in the following text. 

\subsection{Deep learning models}
First, we benchmark various deep learning architectures for QoE modeling. The models compared include 
Multivariate-Transformers~\cite{vaswani2017attention, zerveas2021transformer}, unidirectional and bidirectional 
LSTMs~\cite{hochreiter1997long}, Fully Convolutional Networks (FCNs)~\cite{wang2017time}, 
LSTM-FCN hybrids~\cite{karim2017lstm}, ResNet~\cite{wang2017time}, and 
InceptionTime~\cite{ismail2020inceptiontime}.
Table~\ref{tab:deepperformance} presents the performance of these models on test data, with 
the best Root Mean Square Error (RMSE) values in bold. To optimize hyperparameters, we 
iterated over various combinations of loss rates, number of layers, dropout rates, and other 
parameters. The results shown correspond to the optimized hyperparameters.

Our evaluation shows that, on these datasets, LSTM achieves the best performance among the compared models. 
While Transformers and FCNs perform comparably. Nevertheless, these models have a high count of learnable parameters due to their complex architecture.

\begin{table}
    \centering
    \footnotesize
        \begin{tabular}{|l|c|c|c|}
            \hline
            \textbf{Model} &\textbf{Parameters} &\textbf{LIVE-Netflix-II} & \textbf{WaterlooIII} \\ 
            &\textbf{(both datasets)}&\textbf{RMSE}&\textbf{RMSE}\\
            \hline 
            Transformer & 3350017, 3347201 & 0.2051 & 0.4274 \\
            LSTM-FCN&315973, 311973&0.2046&0.4329\\
            FCN&269569, 268673&0.2066&0.4152\\
            ResNet&480897, 481409&0.2017&0.4986\\
            InceptionTime&456129, 455937&0.2116&0.4904\\
            LSTM bi-dir. &328201, 328201&{\bf 0.1850}&0.4107\\
            LSTM uni-dir. &124101, 204901&0.1865&{\bf 0.3889}\\
            \hline
        \end{tabular}
    \caption{Test performance for deep learning models.}
    \label{tab:deepperformance}
\end{table}

\begin{table}
    \centering
    \footnotesize
        \begin{tabular}{|l|c|c|c|}
            \hline
            \textbf{Model} &\textbf{Parameters} &\textbf{LIVE-Netflix-II} & \textbf{WaterlooIII} \\ 
            &\textbf{(both datasets)}&\textbf{RMSE}&\textbf{RMSE}\\
            \hline 
            
            TSKAN&120 & {\bf0.2169} & {\bf0.4716}\\
            DT (as in \cite{wehner2023explainable})& depth=5& 0.6020&0.5909\\
            LR& 6 & 0.5092&0.6695\\
            LASSO& 6 & 0.5461&0.6587\\
            EBM (as in \cite{wehner2023explainable})&7875 & 0.4914&0.4722\\
            NAM (as in \cite{wehner2023explainable}) &11361 & 0.5480&0.5748\\
            \hline
        \end{tabular}
    \caption{Test performance for interpretable models. 
The \textit{Parameters} column indicates: model depth for DT and number of trainable parameters for TSKAN, LR, LASSO, NAM, and EBM.}
    
    \label{tab:tskanperformance}
    \vspace{-2em}
\end{table}

\subsection{Interpretable models}
To benchmark the interpretable models, we compare their performance against TSKAN. TSKAN was trained using $F=1$ (components 0 and 1) and top-$k = 10$. We evaluate other recent explainable QoE modeling works~\cite{wehner2023explainable}, as well as classic inherently interpretable models like 
Decision Trees (DT), Linear Regression (LR), Least Absolute Shrinkage and Selection Operator (LASSO), etc.  Additionally, we include 
recent interpretable models such as Explainable Boosting Machine (EBM)~\cite{nori2019interpretml} and Neural Additive Models (NAM)~\cite{agarwal2021neural}, which were also studied in~\cite{wehner2023explainable}. For these models, we utilize the 
expert-crafted features provided in~\cite{wehner2023explainable}.

The performance of these models on test data is presented in Table~\ref{tab:tskanperformance}. Note that their low number of learnable parameters offers a distinct advantage.
TSKAN achieves the best results. Among other interpretable models, EBM comes second in terms of performance. While TSKAN does not surpass the deep learning models in terms of accuracy, its performance is remarkably close, indicating that it has successfully captured the complex relationships 
between input time series data and predicted MOS scores.
The fact that TSKAN outperforms other interpretable models while maintaining its 
interpretability highlights the benefits of using a KAN architecture and frequency domain 
features. The results also suggest that there is still room for improvement, particularly in 
terms of closing the gap between our approach and deep learning models. Future work could 
focus on incorporating additional techniques to further enhance its 
performance.

TSKAN's interpretable nature can be seen in Figure~\ref{fig:allfun}, which 
shows the trained activations. The output on y-axis varies as a function of each feature on x-axis, 
indicating their contributions to predict the MOS score.

TSKAN employs a 1-layer KAN architecture, as depicted in Figure~\ref{fig:kan}. This simple 
design sums the outputs of different activations to produce the final value, which is the 
predicted MOS score. We justify our choice of using 1 layer and selecting best features by 
considering that increasing the number of layers or features can compromise the model's 
interpretability.

\subsection{Interpreting the trained model}
Before interpreting the model, let us understand the frequency domain features. 
$M_v(0)$ is the DC component of feature $v$, computed as the sum of all values in its time series. 
$M_v(1)$ represents the first frequency component, whose wavelength equals the length of the time series (Figure~\ref{fig:phase}). $M_v(1)$ near 0 will mean flat curve with no changes, whereas higher values will mean changing pattern over time.
The corresponding phase $\phi_v(1)$ captures temporal shifts in the waveform, allowing the model to reflect the influence of recent or past changes in the feature values. Such shifts are illustrated in Figure~\ref{fig:phase} with different phases. For example shift of 1.57 radians (or $\pi/2$) means that the high values of the given feature occurred at the end in the time series. This can capture the position of ups and downs of a given feature.

Now lets examine the activations. They provide insights into the relationships between 
the input features and the predicted MOS score.
The activation for $M_{stalling}(0)$ reveals that MOS decreases with increasing sum of 
stalling durations. This makes sense, as prolonged periods of stalling can be frustrating for 
users. In contrast, $M_{stalling}(1)$ shows a positive impact, suggesting that users prefer low number of stalling events with high duration rather than regular, spread-out, stalling events with moderate durations.

The activation for $M_{bitrate}(0)$ exhibits unusual behavior, with its contribution to the 
output fluctuating as bitrate increases.  Its impact is 
relatively small, ranging from -0.04 to 0.04. This may indicate that this feature is not crucial 
for predicting MOS and could be removed from the final model. Such features may be pruned during model debugging. 

Notably, chunksize, related to scene cuts, has a significant positive contribution to the output, with increasing 
chunk duration leading to a negative impact on MOS. This could be due to the fact that larger 
chunk duration indicates lower motion or fewer scene cuts, which may make compression artifacts more noticeable since they persist for longer periods. The phase of the first frequency component 
of chunksize also plays a role, indicating that if motion slowed down at the end (i.e. higher chunk durations) then users did not notice the distortion as compared to when it happened in the beginning or in the middle.

As expected, increasing sum of QP leads to negative impact, as higher 
compression levels generally result in poorer video quality. The phase of QP's first frequency component suggests that when QP or compression is low in the middle of the video sequence then users have better experience as compared to when QP is higher in the middle, which happens when phase either decreases or increases.

The framerate feature shows strange pattern. MOS estimation suddenly drops when frame rate goes from 24 to 25 and then it increases. Such features may be removed as they show strange behavior. Such things could be part of model debugging and refinement, which in turn are enabled by TSKAN. We leave this for future work as we currently focus on TSKAN design and performance.
Finally, higher resolutions consistently result in better QoE, as expected.

\section{Conclusions}
\label{sec:conclusion}
We proposed TSKAN, an interpretable ML model for predicting QoE. By incorporating 
frequency domain features and using a 1-layer KAN architecture, TSKAN captures the 
complex relationships between input time series data and MOS scores. From interpretable activations, we gained insights into the individual contributions of each feature to the output, 
revealing interesting patterns. This allows for more informed decision-making in video streaming applications.
Our results show that TSKAN outperforms other interpretable models. Overall, TSKAN's interpretability offers a promising approach for predicting QoE, enabling service providers to design optimized video streaming systems that cater to users' preferences and expectations.
In future, we would like to test TSKAN on timeseries datasets from different domains.


\bibliographystyle{IEEEtran}
\bibliography{biblio}

\end{document}